\documentclass{article}

\PassOptionsToPackage{numbers,compress}{natbib}

\usepackage[main, final]{neurips_2026}

\usepackage[utf8]{inputenc} 
\usepackage[T1]{fontenc}    
\usepackage{url}            

\usepackage{booktabs}       
\usepackage{amsfonts}       
\usepackage{nicefrac}       
\usepackage{microtype}      
\usepackage{xcolor}         
\usepackage{algpseudocode}
\usepackage[colorlinks=true, citecolor=blue, linkcolor=blue, urlcolor=blue]{hyperref}
\usepackage{microtype}
\usepackage{graphicx}
\usepackage{subfigure} 
\usepackage{bm}
\usepackage{comment}
\usepackage{amsfonts}
\usepackage{amsmath}
\usepackage{balance}
\usepackage{color}
\usepackage{wrapfig}
\usepackage{listings} 
\usepackage{xcolor} 
\usepackage{arydshln}
\usepackage{cases} 
\usepackage{booktabs}
\usepackage{algorithm}
\usepackage{threeparttable}
\usepackage{multicol,multirow}  
\usepackage{amssymb}
\usepackage{mathtools, nccmath}
\usepackage{pifont}
\usepackage{amsthm}

\title{Nonlinear Direct Feedback Alignment for Scalable Backpropagation-Free Training}

\author{
Wu Lin$^1$, Yan Fan$^1$, Chenxiang Ma$^{2}$\thanks{Corresponding author.}~, Yinglan Feng$^{3,4}$, \\
\textbf{Jia Wang$^1$, Ka-Chun Wong$^5$, Qiuzhen Lin$^1$}\\
$^1$ College of Computer Science and Software Engineering, Shenzhen University, \\
$^2$ School of Artificial Intelligence, Tianjin University, \\
$^3$ School of Artificial Intelligence, Shenzhen University, \\
$^4$ Intelligent Computing Center, Shenzhen University, \\
$^5$ Department of Computer Science, City University of Hong Kong \\
\small{\texttt{\{linwu, yinglfeng, jia.wang, qiuzhlin\}@szu.edu.cn},
\texttt{2500101043@mails.szu.edu.cn},} \\
\small{\texttt{machenxiang@tju.edu.cn},
\texttt{kc.w@cityu.edu.hk}}
}

\begin{document}

\maketitle

\begin{abstract}
Backpropagation (BP) remains the predominant algorithm for training deep neural networks, but it is biologically implausible and suffers from the backward locking problem, constraining parallelization across layers.
Direct Feedback Alignment~(DFA), which propagates the global output error directly to each hidden layer via fixed feedback matrices, has emerged as a promising alternative for addressing these limitations. However, a significant performance gap persists between DFA and BP, which we attribute to the limited ability of fixed feedback matrices to capture the nonlinear transformation from hidden representations to the network output. To address this limitation, we propose Nonlinear Direct Feedback Alignment (NDFA), which constructs each hidden-layer feedback using a lightweight nonlinear surrogate subnetwork. To enhance the effectiveness of the feedback throughout training, we further propose a learning strategy to update the surrogate subnetwork, thereby better approximating the underlying nonlinear transformation. Extensive experiments across residual networks and vision transformers on CIFAR-10, CIFAR-100, SVHN, and ImageNet datasets demonstrate that NDFA consistently outperforms existing DFA-based methods and substantially narrows the performance gap to BP, while maintaining efficient parallel training. The code is available at https://github.com/YanFan321/NDFA.

\end{abstract}

\section{Introduction}
\label{sec:Introduction}
Backpropagation (BP)~\cite{BP} has long been the dominant algorithm for training deep neural networks, underpinning the success of deep learning across a wide range of applications~\cite{lecun2015deep,wang2024pssdtransformer,wang2025spikeretinexformer,liang2026spikehtr,liang2026smm}. Despite its remarkable effectiveness, BP relies on layer-by-layer gradient propagation, forcing each hidden layer to wait for gradients from subsequent layers before updating its parameters. This issue, known as the backward locking problem~\cite{jaderberg2017decoupled}, hinders efficient parallelization of the training process. Additionally, this raises concerns about the biological plausibility of BP~\cite{lillicrap2020backpropagation,richards2019deep}, as synaptic plasticity in the brain is believed to operate locally and in parallel, without coordinated signals across distant layers~\cite{hebb2005organization,caporale2008spike}.

Direct feedback alignment (DFA)~\cite{nokland2016direct} has emerged as a promising alternative for training deep neural networks. Unlike BP, DFA propagates error signals directly from the output layer to each hidden layer through fixed random feedback matrices, eliminating the need for layer-by-layer gradient propagation. Consequently, DFA addresses the backward locking problem, as each layer can be updated independently, enabling efficient parallelization of the training process. Additionally, DFA is considered more biologically plausible because it supports local and parallel parameter updates while the global output error resembles 
neuromodulators that influence local synaptic plasticity~\cite{roelfsema2018control,clark2021credit,kusmierz2017learning}. Furthermore, DFA is well-suited to emerging non-von Neumann hardware, where BP is challenging to implement~\cite{nakajima2022physical,filipovich2022silicon,lee2020spike,ren2025pipelined}. For example, DFA can be efficiently implemented via multiple scattering of light in optical processing units, achieving substantially higher throughput and energy efficiency than BP on conventional hardware~\cite{wang2025streamlinedopticaltraininglargescale}. 

Although DFA can be applied to a variety of modern deep learning tasks and architectures~\cite{2020DFAScales,refinetti2021align,lee2026enabling}, its performance still lags significantly behind BP~\cite{bartunov2018assessing,crafton2019direct,launay2019principled}, particularly in large-scale networks~\cite{moskovitz2019feedbackalignmentdeepconvolutional}. To narrow this gap, existing efforts have primarily focused on improving how the feedback matrices in DFA are updated. For instance, Direct Kolen–Pollack (DKP)~\cite{dkp} updates the feedback matrices to align with the corresponding forward weights using the outer product of locally available hidden representations and the global output error. Recent approaches, such as Forward Direct Feedback Alignment (FDFA)~\cite{fdfa} and Gradient-Aligned Projected Error~(GrAPE)~\cite{caillon2026scaling}, leverage forward-mode Jacobian–vector products to refine the feedback matrices. However, these methods still fail to match the accuracy of BP in deep networks.

In this paper, we first analyze the feedback mismatch between DFA and BP. For each hidden layer, the feedback in BP is determined by a nonlinear subnetwork transforming the hidden representation to the network output. In contrast, DFA uses a fixed feedback matrix, which corresponds to a linear transformation. We argue that this structural mismatch between the two feedback mechanisms is a key source of the performance gap and cannot be resolved by simply updating the feedback matrices. 

\begin{figure}[!t]
	\centering
	\includegraphics[width=0.95\textwidth]{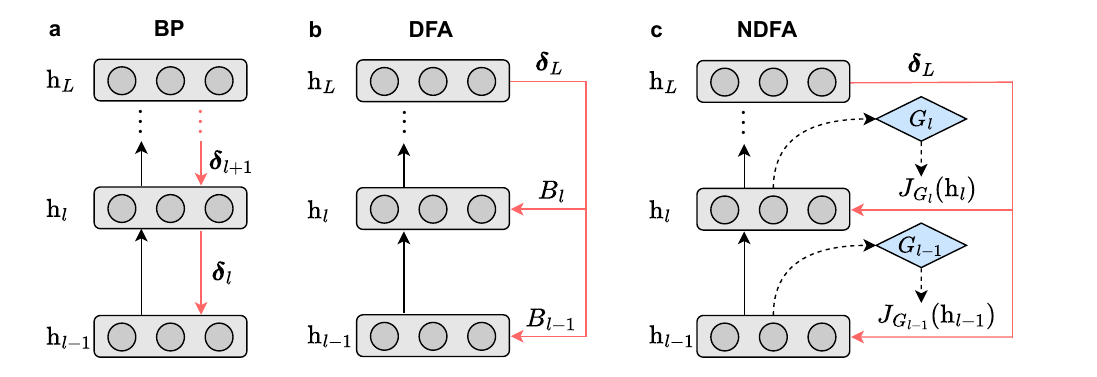}
    \vspace{-2mm}
    \caption{Comparison of (a) BP, (b) DFA, and (c) the proposed NDFA method. BP propagates errors sequentially from the top to bottom layers, leading to the backward locking problem. DFA directly projects the global output error $\delta_L$ to each hidden layer $l$ using fixed random feedback matrices $B_l$. In contrast, NDFA constructs representation-dependent feedback through learnable nonlinear surrogate subnetworks $G_l$.}
    \label{fig:method}
\end{figure}

To address this limitation, we propose Nonlinear Direct Feedback Alignment (NDFA), which introduces a lightweight nonlinear surrogate subnetwork to construct feedback for each hidden layer, as illustrated in Figure~\ref{fig:method}. This surrogate serves as an approximate transformation from the hidden representation to the network output, enabling the feedback to capture representation-dependent nonlinear structure. To encourage the surrogate to better approximate the true transformation, we optimize it using the same target as the true subnetwork. The feedback derived from the surrogate is therefore continuously changed throughout training, allowing it to more effectively guide the learning of each hidden layer. We extensively evaluate the proposed NDFA method on different image classification datasets, including CIFAR-10~\cite{krizhevsky2009learning}, CIFAR-100~\cite{krizhevsky2009learning}, SVHN~\cite{netzer2011reading},
and ImageNet~\cite{deng2009imagenet}, using convolutional and transformer-based architectures, including AlexNet~\cite{krizhevsky2012imagenet}, ResNet-32~\cite{he2016deep}, and ViT-Tiny~\cite{dosovitskiy2020image}. We further evaluate the neural data similarity of NDFA on the Mante 2013 dataset~\cite{mante2013context}.
Our key contributions are summarized as follows:
\begin{itemize}
    \item We identify a key limitation of DFA: its feedback matrices fail to capture the nonlinear transformations from hidden representations to the network output.
    \item We propose NDFA, which leverages a lightweight nonlinear surrogate subnetwork to construct feedback for each hidden layer, thereby addressing the identified limitation. We further propose a learning strategy to update the feedback during training. 
    \item 
    Extensive experiments across diverse datasets and architectures demonstrate that NDFA consistently outperforms existing DFA-based methods, substantially narrows the gap to BP while preserving parallel training efficiency, and achieves strong neural data similarity.
\end{itemize}

\section{Related work}
\label{sec:Related Work}
A growing body of research has explored alternatives to BP~\cite{lillicrap2020backpropagation}. One line of work focuses on addressing the biological implausibility of BP, particularly its reliance on exactly symmetric forward and backward weights~\cite{grossberg1987competitive,crick1989recent}. Feedback Alignment~\cite{lillicrap2016random} shows that propagating error signals through random feedback connections instead of symmetric weights can still support effective learning~\cite{akrout2019deep}. Other approaches avoid exact gradient propagation by directly projecting data labels to hidden layers~\cite{frenkel2021learning} or by propagating layer-wise targets~\cite{lee2015difference,ernoult2022towards}. Despite their biological motivation, these methods struggle to match the performance of BP on challenging tasks and deep architectures~\cite{bartunov2018assessing}.

In addition to biological plausibility, another line of BP-free research promotes parallel learning across layers. Decoupled Neural Interfaces~\cite{jaderberg2017decoupled} introduce synthetic gradients that allow each layer to update independently. Similarly, SoftHebb~\cite{journ2023hebbian} employs local Hebbian learning within soft winner-take-all networks, while other local learning approaches~\cite{belilovsky2019greedy,nokland2019training,ell,stdl} train each hidden layer using an auxiliary network and a local objective. Although these methods improve training parallelism, the lack of cross-layer coordination and global objective guidance often limits their overall performance~\cite{auglocal}.

Recent studies have attempted to eliminate backward passes through forward-only learning~\cite{pepita,srinivasan2024forward}. Present the Error to Perturb the Input To modulate Activity (PEPITA)~\cite{pepita} uses an error-driven extra forward pass, while Forward-Forward~\cite{hinton2022forward} replaces BP with two forward passes over positive and negative samples and trains networks using local goodness functions. Forward gradient methods~\cite{baydin2022gradientsbackpropagation,ren2023scaling,fournier2023can}, in contrast, estimate gradients through directional derivatives computed via forward-mode automatic differentiation. While these approaches offer improved memory efficiency, they often suffer from high-variance gradient estimates and noticeable performance gaps compared with BP.

\section{Background}
\label{sec:Background}
\subsection{Backpropagation (BP)}
\label{sec:bp}
Considering a fully connected neural network with $L$ layers, an input $\mathbf{h}_0$ is propagated sequentially through the network. For each layer $l \in \{1,\ldots,L\}$, the forward pass is given by:
\begin{equation}
\mathbf{a}_l = \mathbf{W}_l \mathbf{h}_{l-1} + \mathbf{b}_l, \qquad
\mathbf{h}_l = \sigma(\mathbf{a}_l),
\end{equation}
where $\mathbf{W}_l\!\in\!\mathbb{R}^{d_l\times d_{l-1}}$ and $\mathbf{b}_l\in\mathbb{R}^{d_{l}}$ denote the weight and bias at layer~$l$, respectively. $\mathbf{a}_l\!\in\!\mathbb{R}^{d_{l}}$ represents the pre-activation, $\sigma(\cdot)$ denotes a nonlinear activation function, and $\mathbf{h}_l\in\mathbb{R}^{d_{l}}$ is the layer's output representation. The network produces a prediction $\hat{\mathbf{y}}\!=\!\mathbf{a}_L\in\mathbb{R}^{d_{y}}$ in the output layer, and a loss function $\mathcal{L}(\hat{\mathbf{y}}, \mathbf{y})$ is then used to measure the difference between the prediction and the target $\mathbf{y}$.

The BP algorithm~\cite{BP} computes the gradients of the loss function with respect to network parameters by sequentially propagating error signals from the output layer to each hidden layer. Let $\boldsymbol{\delta}_l\!=\!\frac{\partial \mathcal{L}}{\partial \mathbf{a}_l}$
denote the hidden-layer error at layer $l$. The global output error, $\boldsymbol{\delta}_L\!=\!\frac{\partial \mathcal{L}}{\partial \mathbf{a}_L}\in\mathbb{R}^{d_{y}}$, can be directly computed by the loss function. For hidden layers, the errors are propagated backward according to:
\begin{equation}
\boldsymbol{\delta}^{\mathrm{BP}}_l = \left(\mathbf{W}_{l+1}^\top \boldsymbol{\delta}^{\mathrm{BP}}_{l+1}\right) \odot \sigma'(\mathbf{a}_l), \quad l = L-1, \ldots, 1,
\label{eq:bp_error}
\end{equation}
where $\odot$ denotes the element-wise product and $\sigma'(\cdot)$ is the derivative of the activation function.

Using these errors, the gradients with respect to weight matrices of hidden layers are computed as:
\begin{equation}
\frac{\partial \mathcal{L}}{\partial \mathbf{W}_l}
= \boldsymbol{\delta}^{\mathrm{BP}}_l \mathbf{h}_{l-1}^\top, 
\qquad l = 1, \ldots, L.
\label{eq:weight_update}
\end{equation}

While BP is the most widely adopted learning algorithm, it suffers from the backward locking problem~\cite{jaderberg2017decoupled}, where the computation of weight gradients at a hidden layer must wait until errors from all downstream layers have been computed. This dependency prevents parallelization across layers.

\subsection{Direct feedback alignment (DFA)}
DFA~\cite{nokland2016direct} is an alternative learning algorithm that addresses the backward locking problem in BP. Instead of propagating errors sequentially on a layer-by-layer basis, DFA computes the error at each hidden layer by directly projecting the global output error $\boldsymbol{\delta}_L$ using fixed random feedback matrices. Formally, let $\mathbf{B}_l \in \mathbb{R}^{d_l \times d_y}$ denote a fixed feedback matrix at layer $l$, and the error gradient is computed as:
\begin{equation}
\boldsymbol{\delta}^{\mathrm{DFA}}_l = \left(\mathbf{B}_l \boldsymbol{\delta}_L \right) \odot \sigma'(\mathbf{a}_l), \qquad l = 1, \ldots, L-1.
\end{equation}
The corresponding weight gradients are then computed in the same manner as in Eq.~\eqref{eq:weight_update}. By directly broadcasting the global output error to all hidden layers, DFA eliminates the dependency on error propagation across layers, thereby enabling the hidden layers to be updated in parallel.

\section{Method}
\label{sec:Method}
We first analyze the feedback mismatch between DFA and BP in Section~\ref{subsec:feedback_mismatch}, showing that the feedback matrices in DFA are representation-independent and limited in capturing the nonlinear transformations from hidden representations to the network output. To address these limitations, we propose nonlinear direct feedback alignment (NDFA), which constructs each hidden-layer feedback via a learnable nonlinear surrogate subnetwork.
Section~\ref{subsec:rd_feedback} details the surrogate subnetwork and the resulting feedback, and Section~\ref{subsec:learning_feedback} presents the learning strategy for the surrogate subnetwork.

\subsection{Feedback mismatch between DFA and BP}
\label{subsec:feedback_mismatch}
We begin by analyzing the mismatch between DFA~\cite{nokland2016direct} and BP~\cite{BP} in projecting the global output error to each hidden layer, which is referred to as feedback. For each hidden layer~$l$, let $F_l:\mathbb{R}^{d_l} \rightarrow \mathbb{R}^{d_y}$ denote the downstream subnetwork induced by all subsequent layers, which transforms the hidden representation~$\mathbf{h}_l$ into the network output $\hat{\mathbf{y}}\!=\!F_l(\mathbf{h}_l)$. The hidden-layer errors computed by BP in Eq.~\eqref{eq:bp_error} can be reformulated as: 
\begin{equation}
\boldsymbol{\delta}_l^{\mathrm{BP}} = \left(J_{F_l}(\mathbf{h}_l) \,\boldsymbol{\delta}_L \right) \odot \sigma'(\mathbf{a}_l), \quad l = L-1, \ldots, 1,
\label{eq:bp-error}
\end{equation}
which propagates the global output error $\boldsymbol{\delta}_L$ through the Jacobian of the downstream subnetwork. In Eq.~\eqref{eq:bp-error}, $J_{F_l}(\mathbf{h}_l) = (\partial F_l(\mathbf{h}_l)/\partial \mathbf{h}_l)^\top \in \mathbb{R}^{d_l \times d_y}$ is the transposed Jacobian of $F_l$ at $\mathbf{h}_l$. The term $J_{F_l}(\mathbf{h}_l)$ acts as the feedback, and it is a local linearization of the nonlinear subnetwork $F_l$ around the current hidden representation $\mathbf{h}_l$. Importantly, the feedback depends dynamically on both $\mathbf{h}_l$ and $F_l$.  DFA departs from
this behavior by replacing the Jacobian with a fixed
random matrix $\mathbf{B}_l \in \mathbb{R}^{d_l \times d_y}$ that is
independent of both $\mathbf{h}_l$ and $F_l$.

Additionally, the feedback $\mathbf{B}_l$ in DFA implicitly induces a linear transformation~$\mathbf{h}_l \mapsto \mathbf{B}_l^\top
\mathbf{h}_l$ from the hidden representation space to the network output space. 
In contrast, the downstream subnetwork~$F_l$ in BP is highly nonlinear, as it consists of multiple layers and nonlinear activations. This mismatch between a simple linear transformation and a complex nonlinear subnetwork imposes an intrinsic limitation on DFA's ability to approximate BP. As network depth increases, the increasing complexity of $F_l$ amplifies this mismatch. 

The analysis of feedback mismatch between DFA and BP suggests two desirable properties for effective feedback. First, the feedback should be
\emph{representation-dependent}, allowing the hidden-layer error to reflect the state of the network at each hidden representation. Second, its induced transformation should be \emph{nonlinear}, enabling it to better approximate $F_l$.

\subsection{Representation-dependent feedback via nonlinear surrogate subnetworks}
\label{subsec:rd_feedback}
Guided by the above analysis, we aim to design the feedback that is representation-dependent and induced by a nonlinear transformation. 
Intuitively, this requires approximating the downstream subnetwork with a lightweight model, so that the feedback can vary with hidden representations without incurring excessive computational cost.

To formalize this idea, we generalize the feedback from a static matrix to a function of hidden representations. Specifically, for each hidden layer $l$, we introduce a function
$
A_l : \mathbb{R}^{d_l} \rightarrow \mathbb{R}^{d_l \times d_y},
$
which produces the feedback $A_l(\mathbf{h}_l)$ dependent on $\mathbf{h}_l$. This formulation unifies both BP and DFA as special cases: BP corresponds to the ideal but intractable choice $A_l(\mathbf{h}_l) = J_{F_l}(\mathbf{h}_l)$, while DFA corresponds to the degenerate static choice $A_l(\mathbf{h}_l) \equiv \mathbf{B}_l$. Our goal is to design tractable feedback that lies between these two extremes.

To this end, we introduce a nonlinear surrogate subnetwork that approximates the true downstream subnetwork, and construct feedback from its Jacobian. Concretely, we associate each hidden layer $l$ with a lightweight surrogate subnetwork
$
G_l(\mathbf{h}_l;\boldsymbol{\theta}_l): \mathbb{R}^{d_l} \rightarrow \mathbb{R}^{d_y},
$
where $\boldsymbol{\theta}_l$ denotes its parameters.
The feedback is expressed as
$
A_l(\mathbf{h}_l)\!=\!J_{G_l}(\mathbf{h}_l)\!=\!(\partial G_l(\mathbf{h}_l)/\partial \mathbf{h}_l)^\top.
$
The resulting hidden-layer errors in NDFA are computed as follows:
\begin{equation}
\boldsymbol{\delta}^{\mathrm{NDFA}}_l
= \left(J_{G_l}(\mathbf{h}_l) \boldsymbol{\delta}_L\right)
  \odot \sigma'(\mathbf{a}_l), \qquad l = 1, \ldots, L-1.
\end{equation}
To strike a balance between expressiveness and computational efficiency, we instantiate $G_l(\mathbf{h}_l;\boldsymbol{\theta}_l)$ as a multilayer perceptron (MLP), which provides a lightweight surrogate with strong empirical performance, as further discussed in Appendix~\ref{app:ablation_study}: \begin{equation}
G_l(\mathbf{h}_l;\boldsymbol{\theta}_l)
= \mathbf{M}_{l, 2}\,\sigma\!\bigl(\mathbf{M}_{l, 1}\mathbf{h}_l\bigr),
\end{equation}
where $\boldsymbol{\theta}_l = \{\mathbf{M}_{l, 1}\in\mathbb{R}^{r\times d_l},\;
                         \mathbf{M}_{l, 2}\in\mathbb{R}^{d_y\times r}\}$, and $r$ controls the hidden width.

This design yields two key advantages. First, the surrogate subnetwork is nonlinear and better aligns with the structure of the downstream subnetwork. Second, the resulting feedback is representation-dependent, allowing the hidden-layer errors to adapt dynamically to hidden representations.

\subsection{Learning nonlinear surrogate subnetworks for effective feedback}
\label{subsec:learning_feedback}
The effectiveness of the proposed feedback depends on how well the surrogate subnetwork $G_l$ approximates the true subnetwork $F_l$. However, directly matching $G_l$ to $F_l$ is difficult, since $F_l$ is determined by the subsequent layers that are constantly updated during training.

To address this, we train the surrogate $G_l$ to approximate the input-output behavior of the true subnetwork $F_l$ on the hidden representations encountered during training. This is motivated by the fact that, despite its evolving form, the role of $F_l$ is to map hidden representations to the target space. By aligning $G_l$ with this functional behavior, the surrogate could track $F_l$ over the course of training.
 
Formally, for each hidden layer $l$, we define a feedback adaptation objective over the empirical distribution of hidden representations:
\begin{equation}
\mathcal{L}_{\,l}^{\mathrm{fb}}
= \mathbb{E}_{\mathbf{h}_l \sim \mathcal{D}_l}
\bigl[\,\mathcal{L}\bigl(G_l(\mathrm{sg}(\mathbf{h}_l)),\, \mathbf{y}\bigr)\,\bigr],
\label{eq:fb_loss}
\end{equation}
where $\mathcal{D}_l$ denotes the distribution of $\mathbf{h}_l$ on the training data, $\mathcal{L}(\cdot,\cdot)$ is the same loss function as the main network, and $\mathbf{y}$ is the corresponding ground-truth label. The operator~$\mathrm{sg}(\cdot)$ denotes the stop-gradient operation, which prevents gradients from propagating into the main network. The parameters of the surrogate subnetwork are updated as follows:
\begin{equation}
\boldsymbol{\theta}_l \leftarrow \boldsymbol{\theta}_l
-\eta\,\nabla_{\boldsymbol{\theta}_l}\mathcal{L}_{\,l}^{\mathrm{fb}},
\label{eq:theta_update}
\end{equation}
where $\eta$ denotes the learning rate shared by the main network and surrogate subnetworks. Notably, the gradients of $\mathcal{L}_{\,l}^{\mathrm{fb}}$ are only used to update $\boldsymbol{\theta}_l$ and do not affect the parameters of the main network.

It is worth noting that, in NDFA, the feedback computation and surrogate subnetwork update for each hidden layer can be performed in parallel with the forward pass of subsequent layers. Once the forward pass is completed and the global output error becomes available, NDFA uses the precomputed feedback to directly project the global output error into each hidden layer, thereby enabling parallel training across layers. The pseudocode is provided in Appendix Algorithm~\ref{alg:ndfa}.

\begin{table}[!t]
	\begin{minipage}{0.55\textwidth}
    \centering
    \setlength{\tabcolsep}{4pt}
    \caption{Comparison of different methods on CIFAR-10, CIFAR-100, and SVHN using ConvNet.}
    \label{tab:cnn}
    \vspace{-0.5em}
    \begin{tabular}{lccc}
    \toprule
    Method & CIFAR-10 & CIFAR-100 & SVHN \\
    \midrule
    BP & 81.36$\pm 0.32$ & 51.40$\pm 0.25$ & 93.41$\pm 0.12$ \\
    \midrule
    
    DFA & 71.59$\pm 0.77$ & 36.03$\pm 0.95$ & 89.09$\pm 1.15$ \\
    DKP & 73.04$\pm 0.13$ & 37.53$\pm 0.71$ & 88.34$\pm 1.62$ \\
    PEPITA & 56.80$\pm 0.56$ & 22.93$\pm 1.62$ & 78.71$\pm 0.63$ \\
    FDFA & 74.32$\pm 0.45$ & 41.84$\pm 1.00$ & 93.42$\pm 0.05$ \\
    \textbf{NDFA} & \textbf{78.62$\pm 0.46$} & \textbf{47.86$\pm 0.54$} & \textbf{93.49$\pm 0.09$} \\
    \bottomrule
    \end{tabular}
    
    \end{minipage}
    \hfill
    \begin{minipage}{0.42\textwidth}
    \centering
    \setlength{\tabcolsep}{6pt}
    \caption{Comparison of different methods using AlexNet.}
    \label{tab:alexnet}
    \vspace{-0.5em}
    \begin{tabular}{lcc}
    \toprule
    Method & CIFAR-10 & CIFAR-100 \\
    \midrule
    BP &  90.95$\pm 0.23$ & 61.25$\pm 0.11$ \\
    \midrule
    DFA  &  80.68$\pm 0.11$ & 48.00$\pm 0.15$ \\
    DKP &  83.63$\pm 0.19$ & 58.00$\pm 0.10$ \\
    PEPITA  & 53.03$\pm0.16$ & 21.03$\pm0.21$ \\
    FDFA &  84.65$\pm 0.12$ & 60.70$\pm 0.31$ \\
    \textbf{NDFA} & \textbf{87.38$\pm 0.14$} & \textbf{61.74$\pm 0.37$} \\
    \bottomrule
    \end{tabular}

    \end{minipage}

\end{table}

\begin{table*}[!t]
\centering
\caption{Comparison of different methods using ResNet-32. $K$ is the number of parallel modules, where $K=16$ represents the layer-wise setting.}
\label{tab:resnet}
\setlength{\tabcolsep}{4.5pt}

\setlength{\aboverulesep}{0pt}
\setlength{\belowrulesep}{0pt}
\renewcommand{\arraystretch}{1.2} %

\begin{tabular}{l : ccc : ccc}
\toprule
\multirow{2}{*}{Method} & \multicolumn{3}{:c:}{$K=4$} & \multicolumn{3}{c}{$K=16$} \\
\cmidrule{2-7}
 & CIFAR-10 & CIFAR-100 & SVHN & CIFAR-10 & CIFAR-100 & SVHN \\
\midrule
BP ($K{=}1$) & 92.23$\pm$0.22 & 67.98$\pm$0.22 & 96.71$\pm$0.09 & 92.23$\pm$0.22 & 67.98$\pm$0.22 & 96.71$\pm$0.09 \\
\midrule
DFA  & 72.73$\pm$0.31 & 27.17$\pm$0.59 & 90.85$\pm$0.33 & 60.62$\pm$2.19 & 15.24$\pm$0.46 & 80.93$\pm$1.07 \\
DKP  & 63.91$\pm$0.56 & 21.84$\pm$0.78 & 90.02$\pm$0.70 & 51.04$\pm$0.61 & 18.59$\pm$0.77 & 65.92$\pm$1.32 \\
FDFA & 80.63$\pm$0.35 & 49.12$\pm$0.63 & 93.25$\pm$0.32 & 77.36$\pm$0.23 & 48.53$\pm$0.22 & 92.77$\pm$0.83 \\
\textbf{NDFA} & \textbf{85.36}$\pm$0.48 & \textbf{55.38}$\pm$0.38 & \textbf{94.82}$\pm$0.04 & \textbf{81.53}$\pm$0.60 & \textbf{50.45}$\pm$0.25 & \textbf{93.42}$\pm$0.19 \\
\bottomrule
\end{tabular}
\vspace{-1.0em}
\end{table*}

\section{Experiments}
\label{sec:Experiments}


\subsection{Experimental setup}

\textbf{Datasets.}
Our evaluation uses four commonly used image classification datasets, including CIFAR-10~\cite{krizhevsky2009learning}, CIFAR-100~\cite{krizhevsky2009learning}, SVHN~\cite{netzer2011reading}, and ImageNet~\cite{deng2009imagenet}.

\textbf{Architectures.}
We use four representative architectures, including a ConvNet with two convolutional layers and two fully connected layers, AlexNet~\cite{krizhevsky2012imagenet}, ResNet-32~\cite{he2016deep}, and ViT-Tiny~\cite{dosovitskiy2020image}.

\textbf{Baselines.}
We compare NDFA with BP and existing DFA-based methods, including DFA~\cite{nokland2016direct}, DKP~\cite{dkp}, PEPITA~\cite{pepita}, and FDFA~\cite{fdfa}. 
DFA directly broadcasts the global output error to each hidden layer through fixed random feedback matrices. 
DKP updates the feedback matrices using information from the forward pathway. 
PEPITA perturbs the input using the global output error through an extra forward pass and can be viewed as a DFA-like algorithm.
FDFA updates the feedback matrices using forward-mode gradient estimates.

\textbf{Implementation details.} For fair comparisons, we adopt consistent training settings across all methods, except for the learning rate, which is searched for each method.
We compare NDFA with baselines in a module-wise setting (see Appendix~\ref{app:module_partition}), where the network is divided into $K$ modules. In this setting, DFA-based methods directly propagate the global output error to each module while allowing gradient propagation within individual modules. All reported results are averaged over three independent runs. Detailed training settings and module partitions are provided in Appendix~\ref{app:module_partition}.

\subsection{Main results}

\textbf{NDFA consistently outperforms DFA-based baselines across diverse datasets.}
As shown in Tables~\ref{tab:cnn} and~\ref{tab:alexnet}, NDFA achieves the best performance among all DFA-based methods across CIFAR-10, CIFAR-100, and SVHN on both ConvNet and AlexNet, substantially reducing the gap to BP. In particular, for ConvNet, NDFA surpasses FDFA by more than 4\% and 6\% accuracy on CIFAR-10 and CIFAR-100, respectively. These results demonstrate the effectiveness of the proposed NDFA method.


\textbf{NDFA consistently outperforms DFA-based baselines on deep architectures.}
Table~\ref{tab:resnet} reports the test accuracy of ResNet-32 on CIFAR-10, CIFAR-100, and SVHN under different numbers of parallel modules $K$. NDFA consistently achieves the strongest performance among all DFA-based methods across different datasets and parallelization granularities. Notably, its performance remains robust as the number of parallel modules increases, demonstrating the effectiveness of NDFA under more challenging parallel training settings.
We further evaluate NDFA on ViT-Tiny to examine its scalability to transformer architectures. As shown in Table~\ref{tab:vit}, NDFA substantially improves the performance of DFA on both CIFAR-10 and CIFAR-100.
These results demonstrate that NDFA generalizes effectively across different architectures.

\textbf{NDFA scales effectively to ImageNet.}
We further evaluate the scalability of NDFA on the large-scale ImageNet dataset using ResNet-32 and ViT-Tiny. As shown in Table~\ref{tab:imagenet}, NDFA substantially outperforms DFA on both architectures. In particular, NDFA improves the accuracy of ResNet-32 by more than 17\%. The improvement is even more significant for ViT-Tiny, where DFA fails to train effectively. These results demonstrate the strong scalability of NDFA to large-scale datasets

\begin{table}[t]
\centering
\begin{minipage}[t]{0.48\textwidth}
    \centering
    \begin{threeparttable}
    \caption{Comparison of different methods on CIFAR-10/100 using ViT-Tiny.}
    \label{tab:vit}
    \vspace{-0.5mm}
    \setlength{\tabcolsep}{3pt}
    \begin{tabular}{lccc}
    \toprule
    Method & $K$ & CIFAR-10 & CIFAR-100 \\
    \midrule
    BP  & 1 & 92.26$\pm0.23$ & 66.08$\pm0.13$ \\
    \midrule
    DFA & 5 & 28.76$\pm0.63$ & 7.00$\pm0.26$  \\
    \textbf{NDFA} & 5 & \textbf{91.15$\pm0.06$} & \textbf{65.22$\pm0.75$} \\
    \midrule
    DFA & 13 & 24.31$\pm0.65$ & 4.69$\pm0.37$  \\
    \textbf{NDFA} & 13 & \textbf{77.88$\pm0.80$} & \textbf{48.91$\pm0.75$} \\
    \bottomrule
    \end{tabular}
    \end{threeparttable}
\end{minipage}
\hspace{-0.02\textwidth}
\begin{minipage}[t]{0.42\textwidth}
    \centering
    \begin{threeparttable}
    \caption{Results on ImageNet using ResNet-32 and ViT-Tiny.}
    \label{tab:imagenet}
    \vspace{-0.5mm}
    \setlength{\tabcolsep}{4pt}
    \begin{tabular}{lcccc}
    \toprule
    Method & \multicolumn{2}{c}{ResNet-32} & \multicolumn{2}{c}{ViT-Tiny} \\
    \midrule
     & $K$ & Acc. & $K$ & Acc. \\
    \midrule
    BP    & 1 & 69.22 & 1 & 73.58 \\
    \midrule
    DFA   & 3 & 44.20 & 3 & 0.72 \\
    \textbf{NDFA} & 3 & \textbf{61.28} & 3 & \textbf{66.27} \\
    \bottomrule
    \end{tabular}
    \end{threeparttable}
\end{minipage}
\vspace{-0.6em}
\end{table}
    

\begin{figure*}[!t]
\centering
\includegraphics[
    width=0.98\textwidth,
    height=0.8\textheight,
    keepaspectratio
]{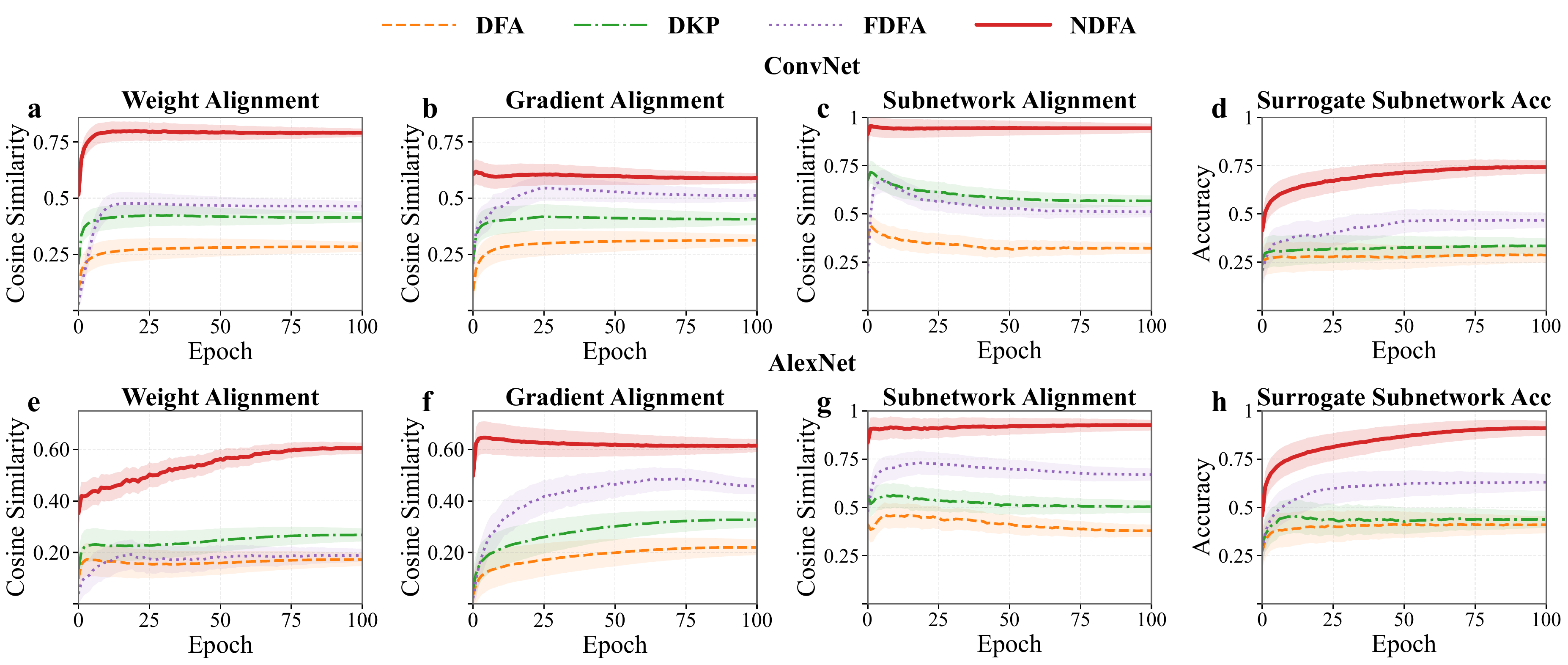}
\vspace{-2mm}
\caption{Comparison of feedback-related analyses between NDFA and other DFA-based methods on CIFAR-10 using ConvNet and AlexNet. Panels (a,e), (b,f), (c,g), and (d,h) show weight alignment, gradient alignment, subnetwork alignment, and surrogate subnetwork accuracy, respectively.
}
\label{fig:alignment}
\end{figure*}

\begin{figure*}[!t]
\centering
\includegraphics[width=0.9 \textwidth]{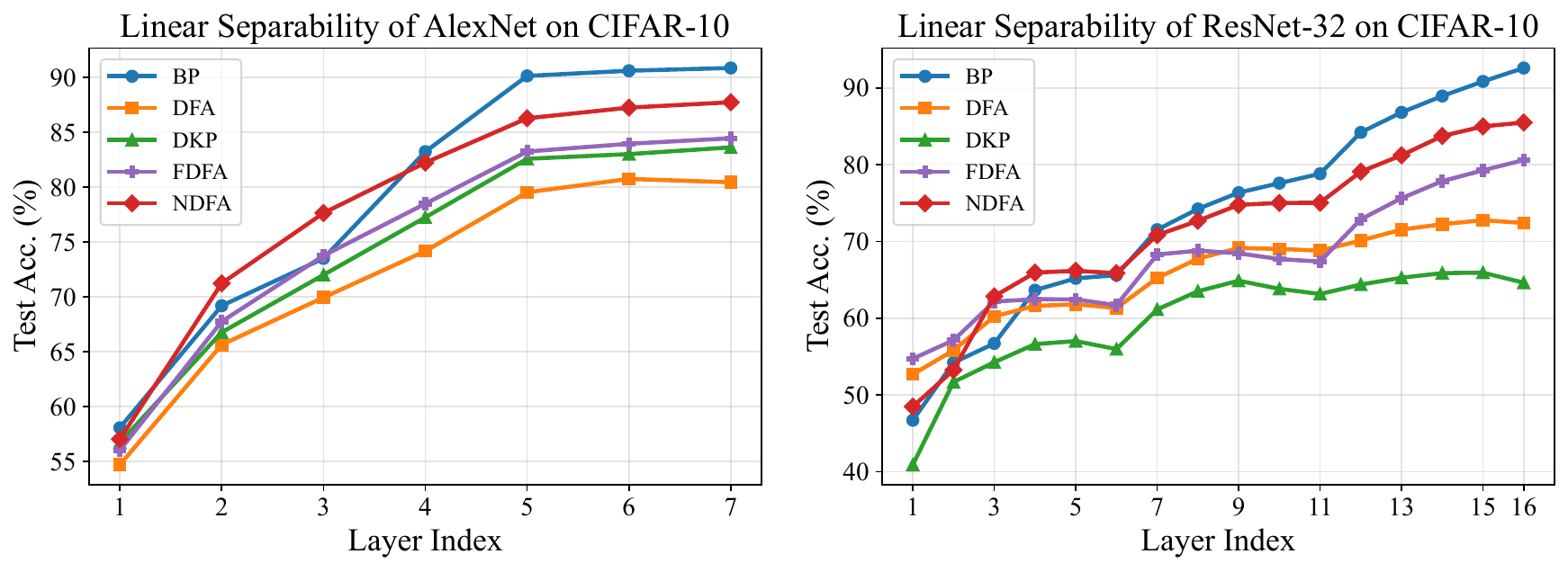}
\vspace{-2mm}
\caption{Comparison of layer-wise linear separability among different methods using AlexNet and ResNet-32 on CIFAR-10.}
\label{fig:linear}
\end{figure*}

\subsection{Analysis of feedback and representations}

\subsubsection{Feedback analysis}

We compare the feedback in NDFA against other DFA-based methods using weight alignment, gradient alignment, subnetwork alignment, and surrogate subnetwork accuracy. Jacobian alignment is provided in Appendix~\ref{app:additional_analysis}. ConvNet and AlexNet are trained on CIFAR-10.

\textbf{Weight alignment} measures the cosine similarity between the feedback and the corresponding forward weight matrix, following previous studies that assume the transformation from a hidden representation to the network output can be approximated by a linear mapping. As shown in Figures~\ref{fig:alignment} (a,e), NDFA achieves substantially stronger weight alignment than existing DFA-based methods. Moreover, the alignment consistently improves throughout training, indicating that the learned feedback in NDFA progressively aligns more closely with the corresponding forward weight.

\textbf{Gradient alignment} measures the cosine similarity between the hidden-layer errors produced by each DFA-based method and BP. As shown in Figure~\ref{fig:alignment} (b, f), NDFA consistently achieves higher gradient alignment with BP than existing DFA-based methods throughout training. This is because NDFA provides a better structural approximation of the Jacobian of BP than baselines. This suggests that the feedback in NDFA produces hidden-layer update directions that more closely approximate those of BP, thereby contributing to its superior performance.

\textbf{Subnetwork alignment} evaluates how well the surrogate subnetwork $G_l$ approximates the true downstream subnetwork $F_l$. Specifically, for the same hidden representation $\mathbf{h}_l$, we compute the cosine similarity between the surrogate output $G_l(\mathbf{h}_l)$ and the true output $F_l(\mathbf{h}_l)$. As shown in Figure~\ref{fig:alignment} (c,g), NDFA achieves the highest subnetwork alignment among all methods. This indicates that the nonlinear surrogate subnetworks in NDFA closely approximate the input-output behavior of the true subnetworks. Note that a more accurate approximation of the downstream subnetwork $F_l$ by its surrogate $G_l$ does not necessarily imply better alignment between their Jacobians.
This is shown in Figure~\ref{fig:alignment} (f,g), where subnetwork alignment gradually increases across epochs, whereas gradient alignment remains stable under NDFA.

\textbf{Surrogate subnetwork accuracy} measures the prediction accuracy of the surrogate output $G_l(\mathbf{h}_l)$ with respect to the ground-truth label $\mathbf{y}$, reflecting how well the surrogate subnetwork learns the predictive function of the corresponding downstream subnetwork.
As shown in Figure~\ref{fig:alignment} (d,h), NDFA achieves substantially higher surrogate subnetwork accuracy than existing DFA-based methods, with the accuracy consistently improving throughout training. 
This indicates that the nonlinear surrogate subnetworks in NDFA can better capture task-relevant downstream transformations, thereby providing more informative feedback signals for learning hidden layers.

\subsubsection{Representation analysis}
We further evaluate the representation quality learned by NDFA using linear probing on CIFAR-10 with AlexNet and ResNet-32. Specifically, we freeze the trained network and train a linear classifier on the representation of each layer. The accuracy of the linear classifier is used to measure the layer-wise linear separability.
Figure~\ref{fig:linear} shows that NDFA consistently learns more discriminative  representations than existing DFA-based methods across both architectures. In particular, NDFA exhibits clear advantages in middle and deep layers, while exhibiting representation trends closer to BP. These results suggest that the feedback in NDFA provides more effective hidden-layer supervision, enabling the network to learn more linearly separable representations.

\subsection{Parallel training efficiency}
Unlike BP, DFA-based methods directly propagate the global output error to each hidden layer without backward locking. Consequently, the gradient computation of different hidden layers can be executed in parallel, potentially leading to significant training acceleration. To evaluate the parallel training efficiency of different methods, we estimate their backward computation cost by using the maximum backward time among all hidden layers. For NDFA, the surrogate subnetworks introduce additional computation overhead. As shown in Appendix Algorithm~\ref{alg:ndfa}, these additional computations can in principle be executed concurrently with the forward pass of subsequent layers. However, to provide a clearer evaluation of the overhead introduced by NDFA, we conservatively include these computations in the measured backward time rather than overlapping them with the forward pass.

Table~\ref{tab:speed} compares the backward computation time of different methods on AlexNet and ResNet-32. Compared with BP, all DFA-based methods achieve substantial reductions in backward computation cost due to their parallel training property. Although NDFA introduces additional overhead compared with DFA, it still achieves significant acceleration over BP and FDFA while substantially outperforming existing DFA-based methods in accuracy. These results demonstrate that NDFA provides a favorable balance between parallel training efficiency and model performance.

\begin{table}[!t]
    \centering
    \begin{minipage}[t]{0.45\textwidth}
    \centering
\caption{Per-iteration backward time (ms) of different methods on CIFAR-10. Values in parentheses denote the speedup relative to BP.
}
    \label{tab:speed}
    \begin{tabular}{lcc}
    \toprule
    Method & AlexNet & ResNet-32 \\
    \midrule
    BP    &  2.34   & 3.45   \\
    \midrule
    DFA   &  0.25 (9.36$\times$)  & 0.24 (14.4$\times$)  \\
    DKP   &  0.33 (7.09$\times$)  & 0.37 (9.32$\times$)  \\
    FDFA  &  1.28 (1.83$\times$)  & 1.11 (3.11$\times$)  \\
    NDFA  &  0.92 (2.54$\times$)  & 0.78 (4.42$\times$)  \\
    \bottomrule
    \end{tabular}
    \vspace{-5mm}
    \end{minipage}%
    \hfill
    \begin{minipage}[t]{0.53\textwidth}
    \centering
    \footnotesize 
    \setlength{\tabcolsep}{2pt} 
    \caption{Ablation study of NDFA using ConvNet and ResNet-32 on CIFAR-10 and CIFAR-100.}
    \label{tab:ablation}
    \vspace{0.72\baselineskip}
    \newcommand{\cmark}{\ding{51}}
    \begin{tabular}{c c c|c c}
    \specialrule{0.08em}{0pt}{0pt}
    Network & Learnable & Nonlinear & CIFAR-10 & CIFAR-100 \\
    \specialrule{0.06em}{0pt}{0pt}
    
    \multirow{4}{*}{ConvNet}
      & \ding{55}     & \ding{55}     & 71.59$\pm0.77$ & 36.03$\pm0.95$ \\
      & \ding{51} & \ding{55}     & 73.17$\pm0.35$ & 39.47$\pm0.67$ \\
      &  \ding{55}    & \ding{51} & 74.60$\pm0.09$ & 41.65$\pm0.15$ \\
      & \ding{51} & \ding{51} & \textbf{78.62$\pm0.46$} & \textbf{47.86$\pm0.54$} \\
    
    \specialrule{0.06em}{0pt}{0pt}
    
    \multirow{4}{*}{ResNet-32}
      & \ding{55}    & \ding{55}     & 72.73$\pm0.31$ & 27.17$\pm0.59$ \\
      & \ding{51} & \ding{55}     & 73.78$\pm0.67$ & 36.40$\pm0.31$ \\
      &  \ding{55}    & \ding{51} & 76.04$\pm0.38$ & 40.10$\pm0.86$ \\
      & \ding{51} & \ding{51} & \textbf{85.36$\pm0.48$} & \textbf{55.38$\pm0.38$} \\
    
    \specialrule{0.08em}{0pt}{0pt}
    \end{tabular}
    \vspace{-0.8em}
    \end{minipage}
\end{table}

\subsection{Ablation study}
We conduct ablation studies to investigate the effectiveness of the two key components of NDFA. Linear feedback is used to denote basic matrix as feedback. Table~\ref{tab:ablation} summarizes the results on ConvNet and ResNet-32 over CIFAR-10 and CIFAR-100. First, introducing learnable feedback consistently improves performance over fixed feedback across both architectures and datasets, highlighting the importance of adapting the feedback to the evolving downstream subnetworks during training. Second, nonlinear feedback substantially outperforms linear feedback regardless of whether the feedback is learnable, demonstrating the importance of modeling nonlinear downstream transformations. Finally, combining both learnable and nonlinear feedback achieves the best performance across all settings, indicating that the two components are complementary for improving feedback quality. Overall, these results demonstrate that both learnable and nonlinear feedback are essential for effective NDFA training.

\subsection{Neural data similarity}

Biologically inspired learning rules are expected not only to achieve strong task performance, but also to produce representations that resemble biological neural recordings. To evaluate the neural data similarity of NDFA, we conduct experiments on the Mante 2013 dataset~\cite{mante2013context}, a widely used neuroscience benchmark containing neural recordings from the prefrontal cortex during a context-dependent decision-making task. Following the experimental setup in~\cite{liu2025can}, we train recurrent neural networks on this dataset using BPTT and NDFA. Implementation details are provided in Appendix~\ref{app:mante}. To quantify neural similarity, we compare the representations of trained models with recorded neural data using Procrustes analysis, where a smaller Procrustes distance indicates stronger alignment with neural recordings. Results are averaged over three independent runs.

\begin{wrapfigure}{r}{0.55\textwidth}
\vspace{-3mm}
    \centering
    \includegraphics[width=0.42\textwidth]{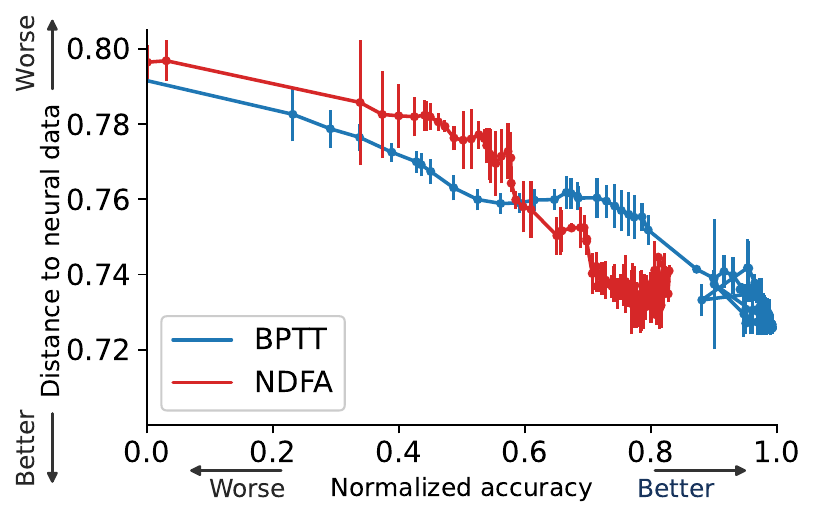}
    \vspace{-2mm}
    \caption{Comparison of neural data similarity between BPTT and NDFA on the Mante 2013 dataset~\cite{mante2013context}.
    }
    \label{fig:Mante}
\end{wrapfigure}

Figure~\ref{fig:Mante} shows the relationship between normalized task accuracy and Procrustes distance throughout training. As task performance improves, both BPTT and NDFA progressively approach the neural data. Notably, in the high-accuracy regime, NDFA consistently achieves lower Procrustes distance than BPTT, indicating that the representations learned by NDFA more closely resemble the recorded neural data. These results suggest that NDFA can achieve competitive task performance while exhibiting stronger neural data similarity.

\section{Conclusion}
\label{sec:Conclusion}

In this paper, we revisit the feedback mismatch between DFA and BP and show that the fixed feedback matrices in DFA are limited in capturing the nonlinear downstream transformations of deep networks. Motivated by this analysis, we propose NDFA, which constructs hidden-layer feedback through learnable nonlinear surrogate subnetworks. Extensive experiments across convolutional networks, residual networks, and vision transformers demonstrate that NDFA consistently outperforms existing DFA-based methods while maintaining parallel training across layers. Further analyses demonstrate that NDFA achieves stronger alignment with BP and learns more discriminative intermediate representations. In addition, experiments on neural data similarity show that NDFA produces representations that more closely resemble biological neural recordings.

\textbf{Limitations.}
Although NDFA substantially narrows the performance gap between DFA and BP, an observable gap to BP still remains on very deep architectures. Future work could investigate more expressive surrogate subnetworks to further improve feedback quality. In addition, NDFA introduces extra computational overhead due to the surrogate subnetworks. While these computations can in principle be executed concurrently with the forward pass, the practical training efficiency of NDFA depends on efficient parallel implementations and hardware support. Exploring asynchronous implementations and system-level acceleration strategies is an important direction for future work. Furthermore, a rigorous theoretical analysis of convergence would provide a deeper understanding of the proposed method, and we leave it for future work. Exploring alternatives to lightweight surrogate subnetworks for constructing the feedback $A_l(\mathbf{h}_l)$ is also a promising future direction.

\section*{Acknowledgments and Disclosure of Funding}

This work was supported in part by the National Natural Science Foundation of China under Grants 62606328, 62602434, 62476170, and 62376163; and in part by the Shenzhen Science and Technology Program under Grant JCYJ20250604181716022. This work is supported by the Intelligent Computing Center of Shenzhen University.

\bibliography{ref}

\bibliographystyle{unsrtnat}

\clearpage
\appendix

\begin{algorithm}[!htb]
\caption{Nonlinear Direct Feedback Alignment (NDFA)}
\label{alg:ndfa}
\begin{algorithmic}[1]
\Require Main network $\{f_l(\cdot;\mathbf{W}_l)\}_{l=1}^{L}$;
surrogate subnetworks
$\{G_l(\cdot;\boldsymbol{\theta}_l)\}_{l=1}^{L-1}$;
learning rate $\eta$

\For{each training iteration}
    \State Sample a mini-batch $(\mathbf{x},\mathbf{y})$
    \State $\mathbf{h}_0 = \mathbf{x}$

    \For{$l=1$ to $L$} \Comment{Forward pass}
        \State $\mathbf{a}_l = f_l(\mathbf{h}_{l-1};\mathbf{W}_l)$
        \If{$l < L$}
            \State $\mathbf{h}_l = \sigma_l(\mathbf{a}_l)$
            \State Compute $\mathcal{L}^{\mathrm{fb}}_l
            =
            \mathcal{L}\!\left(G_l(\mathrm{sg}(\mathbf{h}_l)),\mathbf{y}\right)$
            \hfill $\triangleright$ can run in parallel with the forward pass

            \State Compute $J_{G_l}(\mathbf{h}_l)$ and
            $\nabla_{\boldsymbol{\theta}_l}\mathcal{L}^{\mathrm{fb}}_l$
            \hfill $\triangleright$ can run in parallel with the forward pass
        \Else
            \State $\hat{\mathbf{y}} = \mathbf{a}_L$
        \EndIf
    \EndFor

    \State $\mathcal{L} = \mathcal{L}(\hat{\mathbf{y}},\mathbf{y})$,
    $\boldsymbol{\delta}_L = \nabla_{\hat{\mathbf{y}}}\mathcal{L}$

    \For{$l=1$ to $L$ \textbf{in parallel}}
        \If{$l < L$}

            \State $\boldsymbol{\delta}_l
            =
            \left(
            J_{G_l}(\mathbf{h}_l)
            \boldsymbol{\delta}_L
            \right)
            \odot
            \sigma_l'(\mathbf{a}_l)$

            \State $\nabla_{\mathbf{W}_l}\mathcal{L}
            =
            \boldsymbol{\delta}_l
            \mathbf{h}_{l-1}^{\top}$

            \State $\mathbf{W}_l
            =
            \mathbf{W}_l
            -
            \eta
            \nabla_{\mathbf{W}_l}\mathcal{L}$
            \State $\boldsymbol{\theta}_l
            =
            \boldsymbol{\theta}_l
            -
            \eta
            \nabla_{\boldsymbol{\theta}_l}
            \mathcal{L}^{\mathrm{fb}}_l$
        \Else
            \State $\nabla_{\mathbf{W}_L}\mathcal{L}
            =
            \boldsymbol{\delta}_L
            \mathbf{h}_{L-1}^{\top}$

            \State $\mathbf{W}_L
            =
            \mathbf{W}_L
            -
            \eta
            \nabla_{\mathbf{W}_L}\mathcal{L}$
        \EndIf
    \EndFor
\EndFor
\end{algorithmic}
\end{algorithm}

\section{Implementation details for image classification experiments}
\label{app:hyperparams}

\paragraph{Datasets.} We adopt four image classification datasets, including CIFAR-10~\cite{krizhevsky2009learning}, CIFAR-100~\cite{krizhevsky2009learning}, SVHN~\cite{netzer2011reading}, and ImageNet~\cite{deng2009imagenet}. 
CIFAR-10 is a widely used small-scale natural image classification benchmark consisting of 60,000 color images of size $32 \times 32$. 
It contains 10 classes, with 50,000 training images and 10,000 test images. 
Following standard practice, we use random crop with 4-pixel padding and random horizontal flip for training.
CIFAR-100 has the same image format and data split as CIFAR-10, but contains 100 classes. 
This makes it more challenging due to its finer semantic granularity and higher demand on representation learning.
SVHN is a real-world digit recognition benchmark composed of RGB images cropped from natural street-view scenes. 
The dataset contains $32 \times 32$ images from 10 digit classes. 
Although SVHN has limited semantic diversity, it remains challenging due to cluttered backgrounds, illumination changes, and appearance variations in natural scenes. We adopt the standard split with 73,257 training images and 26,032 test images.
ImageNet is used as the large-scale image classification benchmark. 
It contains 1,000 classes, with approximately 1.2 million training images and 50,000 validation images. Training images are cropped to $224 \times 224$ and randomly horizontally flipped, while validation images are resized and center-cropped for evaluation.

\paragraph{Models.}
\label{app:model}
We evaluate NDFA on a diverse set of architectures, including a ConvNet, AlexNet~\cite{krizhevsky2012imagenet}, ResNet-32~\cite{he2016deep}, and ViT-Tiny~\cite{dosovitskiy2020image}. The ConvNet architecture is 15C5-40C5-MP2-128-10, where 15C5 and 40C5 denote convolutional layers with 15 and 40 output channels and 5×5 kernels, MP2 denotes a 2×2 max-pooling layer, 128 denotes a fully connected hidden layer with 128 units, and 10 denotes the output dimension.  For a convolutional layer, its output feature map is pooled and flattened before being passed to the first linear layer of the corresponding surrogate subnetwork.

\paragraph{Training setups.}
We compare NDFA with BP~\cite{BP} and DFA-based methods, including DFA~\cite{nokland2016direct}, DKP~\cite{dkp}, PEPITA~\cite{pepita}, and FDFA~\cite{fdfa}. All experiments are conducted on 4 NVIDIA RTX A6000 GPUs. The hyperparameters for different methods and architectures are summarized in Table~\ref{tab:hyperparams_all}. Since the optimal learning rate varies across different learning rules, we tune the learning rate separately for each method to ensure fair comparisons.

\paragraph{Module-wise settings.}
\label{app:module_partition}
In module-wise experiments, the network is partitioned into $K$ parallel modules. For ResNet-32 on CIFAR-10 and CIFAR-100, we evaluate $K=4$ and $K=16$. For $K=4$, the network is divided into the convolutional stem and three residual stages, whereas for $K=16$, a finer partition is adopted consisting of the convolutional stem and 15 individual residual blocks.

For ViT-Tiny, we evaluate $K=5$ and $K=13$. For $K=5$, the model is divided into the patch embedding module and four groups of three transformer blocks. For $K=13$, the patch embedding module and each of the 12 transformer blocks are treated as separate modules.

For ImageNet experiments, both ResNet-32 and ViT-Tiny are evaluated with $K=3$, where the backbone is partitioned into three consecutive macro modules to maintain consistent settings across architectures.

\begin{table}[!t]
\centering
\caption{Training hyperparameters used for different methods and architectures.}
\label{tab:hyperparams_all}
\begin{tabular}{llcccc}
\toprule
Architecture & Method & Learning Rate & Optimizer & Batch Size & Epochs \\
\midrule

\multirow{6}{*}{ConvNet}
& BP     & 5e-4 & AdamW & 256 & 100 \\
& DFA    & 5e-4 & AdamW & 256 & 100 \\
& DKP    & 8e-4 & AdamW & 256 & 100 \\
& PEPITA & 5e-2 & SGD   & 256 & 100 \\
& FDFA   & 1e-4 & AdamW & 256 & 100 \\
& NDFA   & 1e-3 & AdamW & 256 & 100 \\

\midrule

\multirow{6}{*}{AlexNet}
& BP     & 5e-4 & AdamW & 256 & 100 \\
& DFA    & 1e-4 & AdamW & 256 & 100 \\
& DKP    & 1e-4 & AdamW & 256 & 100 \\
& PEPITA & 3e-3 & SGD   & 256 & 100 \\
& FDFA   & 1e-4 & AdamW & 256 & 100 \\
& NDFA   & 5e-4 & AdamW & 256 & 100 \\

\midrule

\multirow{5}{*}{ResNet-32}
& BP   & 2e-1  & SGD   & 256 & 200 \\
& DFA  & 5e-4 & AdamW & 256 & 200 \\
& DKP  & 4e-4 & AdamW & 256 & 200 \\
& FDFA & 5e-4 & AdamW & 256 & 200 \\
& NDFA & 1e-3 & AdamW & 256 & 200 \\

\midrule

\multirow{3}{*}{ViT-Tiny}
& BP   & 1e-3 & AdamW & 256 & 300 \\
& DFA  & 1e-3 & AdamW & 256 & 300 \\
& NDFA & 1e-3 & AdamW & 256 & 300 \\

\bottomrule
\end{tabular}
\end{table}

\section{Additional experiments}
\label{app:additional_experiments}

\paragraph{Evaluation on different tasks.}
We evaluate NDFA on a time-series classification dataset (FordA) and a language-modeling dataset (WikiText-2). ViT-Tiny is trained using each evaluated method for 100 epochs, with a batch size of 256 and a learning rate of 0.001. As shown in Table~\ref{tab:cross_task}, NDFA consistently outperforms DFA and achieves performance comparable to BP, demonstrating its effectiveness across different tasks.

\begin{table}[H]
\centering
\caption{Results on FordA and WikiText-2 using ViT-Tiny.}
\label{tab:cross_task}
\setlength{\tabcolsep}{12pt}
\begin{tabular}{lcc}
\toprule
Method & FordA Acc. ($\uparrow$) & WikiText-2 PPL ($\downarrow$) \\
\midrule
BP   & $83.46\pm0.31$ & $84.02\pm0.58$ \\
DFA  & $72.18\pm0.74$ & $229.64\pm4.27$ \\
NDFA & $\mathbf{82.71\pm0.45}$ & $\mathbf{87.96\pm1.17}$ \\
\bottomrule
\end{tabular}
\end{table}

\paragraph{Comparison with different BP-free methods.}
We compare NDFA with Forward-Forward~\cite{hinton2022forward}, Local Greedy Forward Gradient (LG-FG-A)~\cite{ren2023scaling}, Decoupled Neural Interfaces (DNI)~\cite{jaderberg2017decoupled}, and PredSim~\cite{nokland2019training} on CIFAR-10 using the ConvNet. We reproduce Forward-Forward, DNI, and PredSim using their code and compare them with NDFA under a consistent experimental setting, using the same architecture, data preprocessing, training protocol, and hyperparameters. Specifically, all methods are trained for 100 epochs using AdamW with a learning rate of 0.001 and a batch size of 256 using the ConvNet architecture. We also provide the results for LG-FG-A, Forward-Forward, and DNI reported in their respective original papers, as denoted by * in Table~\ref{tab:broad_baselines}. Forward-Forward was originally evaluated using a four-layer fully connected network, while our experiments show that it performs less effectively on convolutional architectures. LG-FG-A relies on its specially designed LocalMixer architecture and is trained for 200 epochs, twice our training budget. Despite these specialized architectures and larger training budgets, NDFA still achieves higher accuracy than these baselines. As shown in Table~\ref{tab:broad_baselines}, NDFA achieves the highest accuracy among all compared methods, outperforming PredSim and the other BP-free baselines.

\begin{table}[H]
\centering
\caption{Comparison with different BP-free methods on CIFAR-10 using ConvNet. Results marked with $*$ are copied from their original papers.}
\label{tab:broad_baselines}
\setlength{\tabcolsep}{9pt}
\begin{tabular}{lc}
\toprule
Method & Accuracy (\%) \\
\midrule
Forward-Forward & $27.68\pm0.21$ \\
Forward-Forward$^*$ & $59.00$ \\
LG-FG-A$^*$ & $69.32$ \\
DNI & $62.68\pm0.10$ \\
DNI$^*$ & $64.86$ \\
PredSim & $76.08\pm0.18$ \\
NDFA & $\mathbf{78.62\pm0.46}$ \\
\bottomrule
\end{tabular}
\vspace{-5mm}
\end{table}

\section{Additional analyses}
\label{app:additional_analysis}

\begin{figure*}[!t]
\centering
\includegraphics[width=0.8\textwidth]{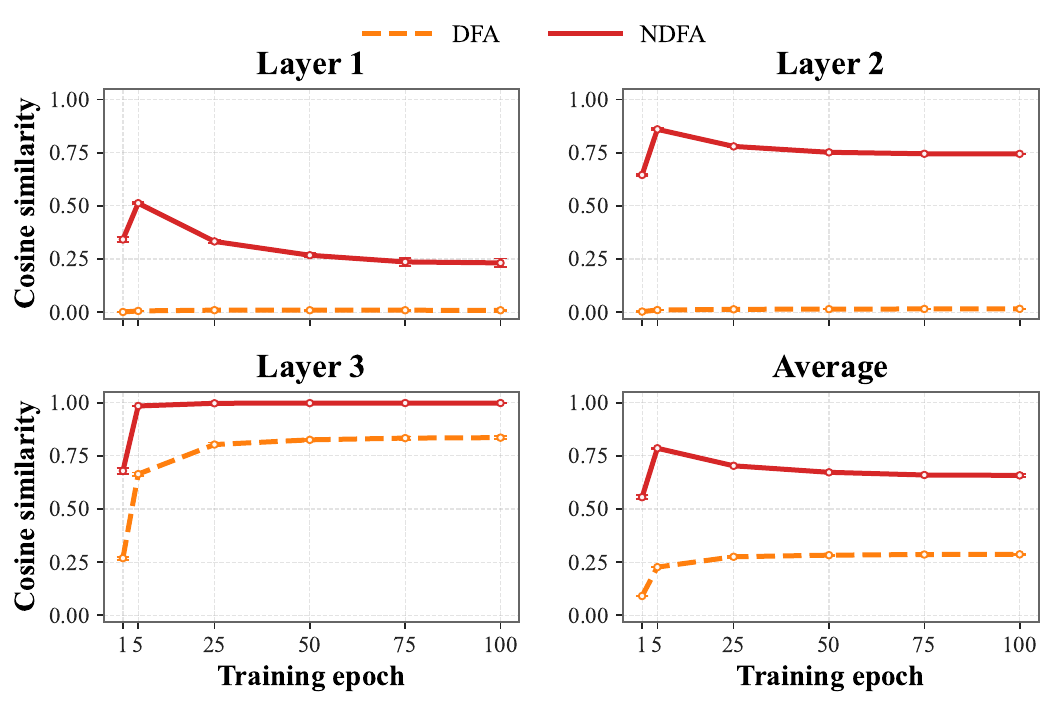}
\vspace{-3mm}
\caption{Comparison of layer-wise Jacobian alignment with BP for NDFA and DFA on CIFAR-10 using a ConvNet.}
\label{fig:jacobian_alignment}
\vspace{-3mm}
\end{figure*}

\paragraph{Jacobian alignment.}
We evaluate the layer-wise cosine similarity between the Jacobians used by NDFA and BP, i.e., $J_{G_l}(\mathbf{h}_l)$ and $J_{F_l}(\mathbf{h}_l)$, on CIFAR-10 using ConvNet. We also compare NDFA with DFA under the same setting. As shown in Figure~\ref{fig:jacobian_alignment}, NDFA consistently achieves substantially higher Jacobian cosine similarity than DFA across different layers and training epochs. This stronger Jacobian alignment helps explain the considerably higher accuracy achieved by NDFA. We also observe that the Jacobian alignment of NDFA increases for layers closer to the network output. This result is expected because earlier layers are followed by deeper and more complex downstream subnetworks, which are more difficult to approximate using lightweight MLP-based surrogate subnetworks. Note that the first-layer alignment for NDFA decreases substantially after reaching an early peak, because the downstream subnetwork associated with the first layer is deeper than other layers. By contrast, the average Jacobian cosine similarity across all layers declines more moderately and remains at a relatively high level. Moreover, both the layer-wise and average Jacobian cosine similarities of NDFA consistently exceed those of DFA throughout training.

\paragraph{Convergence analysis.}
We conduct a convergence analysis by reporting the losses of the main network and three surrogate subnetworks on CIFAR-10 using ConvNet. As shown in Table~\ref{tab:loss_stability}, all losses gradually decrease over the course of training, and their standard deviations remain very small, indicating stable optimization across runs. Moreover, the loss decreases for deeper surrogate subnetwork, suggesting that NDFA can exploit network depth to improve performance.
\begin{table}[H]
\centering
\caption{Convergence analysis on CIFAR-10 using ConvNet.}
\label{tab:loss_stability}
\footnotesize
\setlength{\tabcolsep}{3.5pt}
\renewcommand{\arraystretch}{1.12}
\begin{tabular}{lccccc}
\toprule
Loss & Epoch 1 & Epoch 30 & Epoch 50 & Epoch 80 & Epoch 100 \\
\midrule
Main Network & $1.69\pm0.02$ & $0.74\pm0.02$ & $0.66\pm0.02$ & $0.58\pm0.02$ & $0.57\pm0.02$ \\
1st Surrogate & $3.64\pm0.07$ & $2.26\pm0.06$ & $2.10\pm0.05$ & $1.98\pm0.05$ & $1.96\pm0.05$ \\
2nd Surrogate & $3.51\pm0.04$ & $1.81\pm0.04$ & $1.63\pm0.04$ & $1.48\pm0.04$ & $1.46\pm0.04$ \\
3rd Surrogate & $3.42\pm0.04$ & $1.47\pm0.04$ & $1.28\pm0.04$ & $1.11\pm0.03$ & $1.08\pm0.03$ \\
\bottomrule
\end{tabular}
\end{table}

\paragraph{Influence of learning rate.}
We conduct experiments on CIFAR-10 with ConvNet by varying the learning rate of surrogate subnetworks. As shown in Table~\ref{tab:surrogate_lr}, NDFA remains stable across different learning rates. Moreover, using a larger learning rate for surrogate subnetworks yields improved accuracy, suggesting that further tuning their learning rates could lead to additional accuracy gains.
\begin{table}[H]
\centering
\caption{Results of learning rate (LR) on CIFAR-10 using ConvNet.}
\label{tab:surrogate_lr}
\setlength{\tabcolsep}{10pt}
\begin{tabular}{ccc}
\toprule
LR of Main Net & LR of Surrogate & Accuracy ($\%$) \\
\midrule
$1\times10^{-3}$ & $1\times10^{-3}$ & $78.62\pm0.46$ \\
$1\times10^{-3}$ & $3\times10^{-3}$ & $78.74\pm0.47$ \\
$1\times10^{-3}$ & $5\times10^{-3}$ & $\mathbf{79.34\pm0.50}$ \\
$1\times10^{-3}$ & $5\times10^{-4}$ & $78.58\pm0.42$ \\
$1\times10^{-3}$ & $3\times10^{-4}$ & $78.45\pm0.45$ \\
\bottomrule
\end{tabular}
\end{table}

\paragraph{Effect of nonlinearity in the surrogate subnetworks.}
We conduct additional ablation studies to isolate the effect of nonlinearity in the surrogate subnetworks. Specifically, we compare learnable NDFA with nonlinear surrogate subnetworks against its linear counterpart, while matching the number of parameters between the two variants. We also perform the same comparison for fixed NDFA, again using parameter-matched nonlinear and linear surrogate subnetworks. These experiments are conducted on both CIFAR-10 and CIFAR-100 using ConvNet and ResNet-32.

As shown in Table~\ref{tab:matched_nonlinearity}, both the learnable and fixed versions of NDFA with nonlinear surrogate subnetworks consistently outperform their corresponding linear variants. These results demonstrate the effectiveness of the nonlinearity of the surrogate subnetworks in NDFA.

\begin{table}[H]
\centering
\caption{Effect of nonlinearity in the surrogate subnetworks.}
\label{tab:matched_nonlinearity}
\footnotesize
\setlength{\tabcolsep}{3.6pt}
\renewcommand{\arraystretch}{1.1}
\begin{tabular}{llcc}
\toprule
Network & Method & CIFAR-10 & CIFAR-100 \\
\midrule
\multirow{4}{*}{ConvNet}
& Fixed NDFA w/ Linear Surrogate Subnets & $73.31\pm0.26$ & $40.29\pm0.35$ \\
& Fixed NDFA & $74.60\pm0.09$ & $41.65\pm0.15$ \\
& Learnable NDFA w/ Linear Surrogate Subnets& $76.23\pm0.34$ & $43.42\pm0.58$ \\
& NDFA & $\mathbf{78.62\pm0.46}$ & $\mathbf{47.86\pm0.54}$ \\
\midrule
\multirow{4}{*}{ResNet-32}
& Fixed NDFA w/ Linear Surrogate Subnets & $73.88\pm0.32$ & $37.79\pm0.37$ \\
& Fixed NDFA & $76.04\pm0.38$ & $40.10\pm0.86$ \\
& Learnable NDFA w/ Linear Surrogate Subnets & $81.03\pm0.25$ & $46.64\pm0.47$ \\
& NDFA & $\mathbf{85.36\pm0.48}$ & $\mathbf{55.38\pm0.38}$ \\
\bottomrule
\end{tabular}
\vspace{-5mm}
\end{table}

\section{Implementation details for neural data similarity}
\label{app:mante}
\paragraph{The Mante 2013 dataset~\cite{mante2013context}.} It is a primate neural recording dataset for context-dependent decision-making. In this task, the model must selectively integrate task-relevant sensory inputs according to contextual cues, capturing the core computational requirement of flexible behavior. The dataset enables evaluation of both task performance and the biological plausibility of learned internal representations. In addition, the task involves temporally extended trials consisting of fixation, target presentation, motion or color context cues, a delay period, and a saccadic response.

\paragraph{CTRNN architecture.} 
For the Mante 2013 dataset, we employ a continuous-time recurrent neural network (CTRNN). The model consists of an input-to-hidden projection, recurrent hidden-state dynamics, and a hidden-to-output readout layer. The hidden states evolve continuously over time through recurrent dynamics with nonlinear activations.

\paragraph{Training protocol.}
In this recurrent setting, the BP baseline corresponds to backpropagation through time (BPTT), where errors are propagated backward through the unfolded temporal computation graph. NDFA is extended to the CTRNN by applying the same DFA-based principle to recurrent hidden states. Specifically, at each relevant time step, the global output error is directly projected to the recurrent hidden state through a learned nonlinear surrogate pathway. The surrogate subnetwork maps the current hidden state to the output space and generates a feedback signal, which is then used to compute the hidden-state error for updating the recurrent parameters.

\paragraph{Neural similarity evaluation.}
We follow the evaluation setup in~\cite{liu2025can}. During training, we evaluate both normalized task accuracy and neural similarity at intermediate training checkpoints. Normalized accuracy measures task performance, where higher values indicate better performance. Neural similarity is quantified using the Procrustes distance between the CTRNN hidden representations and recorded neural activity.

Let \(H \in \mathbb{R}^{BT \times N}\) denote the hidden representation of the CTRNN and \(\tilde{H} \in \mathbb{R}^{BT \times N'}\) denote the recorded neural responses, where \(B\) is the number of trials or experimental conditions, \(T\) is the number of time steps, \(N\) is the number of model hidden units, and \(N'\) is the number of recorded neurons. The Procrustes distance is defined as
\[
\theta(H,\tilde{H})
=
\min_{Q \in \mathcal{O}}
\arccos
\left(
\frac{
\langle H_{\phi}, \tilde{H}_{\phi} Q \rangle
}{
\|H_{\phi}\| \|\tilde{H}_{\phi}\|
}
\right),
\]
where \(\mathcal{O}\) denotes the set of orthogonal transformations. A lower Procrustes distance indicates that the learned population activity is more closely aligned with the neural recordings after optimal orthogonal alignment. This evaluation enables comparison between BPTT and NDFA at matched task performance and assesses whether NDFA produces recurrent representations that more closely resemble biological neural activity.

\begin{table}[t]
\centering
\caption{Effects of surrogate subnetwork depth and hidden dimension.}
\label{tab:depth_dimension}
\setlength{\tabcolsep}{6.5pt}
\setlength{\aboverulesep}{0pt}
\setlength{\belowrulesep}{0pt}
\renewcommand{\arraystretch}{1.2}

\begin{tabular}{lcccccc}
\toprule
\multirow{2}{*}{Architecture} & \multicolumn{3}{c}{Dimension=128} & \multicolumn{3}{c}{Surrogate Subnetwork Depth=2} \\
\cmidrule{2-7}
& Depth & CIFAR-10 & CIFAR-100 & Dimension & CIFAR-10 & CIFAR-100 \\
\midrule

\multirow{3}{*}{ConvNet}
& 2 & 78.62$\pm$0.46 & 47.86$\pm$0.54 & 128 & 78.62$\pm$0.46 & 47.86$\pm$0.54 \\
& 3 & 81.30$\pm$0.28 & 49.04$\pm$0.44 & 256 & 79.48$\pm$0.23 & 48.16$\pm$0.47 \\
& 4 & 82.95$\pm$0.22 & 51.11$\pm$0.49 & 512 & 79.82$\pm$0.26 & 48.25$\pm$0.38 \\
\midrule

\multirow{3}{*}{ResNet-32}
& 2 & 85.36$\pm$0.48 & 55.38$\pm$0.38 & 128 & 85.36$\pm$0.48 & 55.38$\pm$0.38 \\
& 3 & 86.76$\pm$0.15 & 58.17$\pm$0.29 & 256 & 85.69$\pm$0.18 & 55.78$\pm$0.35 \\
& 4 & 87.82$\pm$0.14 & 59.35$\pm$0.31 & 512 & 84.81$\pm$0.20 & 55.94$\pm$0.42 \\
\bottomrule
\end{tabular}
\end{table}

\begin{table}[t]
\centering
\caption{Per-iteration backward computation time (ms) of NDFA with different surrogate subnetwork depths $d$ on CIFAR-10 using AlexNet and ResNet-32. Values in parentheses indicate the speedup relative to BP.}
\label{tab:time}
\begin{tabular}{lcc}
\toprule
Method & AlexNet & ResNet-32 \\
\midrule
BP    &  2.34 (1.00$\times$)  & 3.45 (1.00$\times$)   \\
\midrule
NDFA ($d=2$)  &  0.92 (2.54$\times$)  & 0.78 (4.42$\times$)  \\
NDFA ($d=3$)  &  2.23 (1.05$\times$)  & 2.90 (1.19$\times$)  \\
NDFA ($d=4$)  &  3.29 (0.71$\times$)  & 4.87 (0.71$\times$)  \\
\bottomrule
\end{tabular}
\vspace{-5mm}
\end{table}

\section{Influence of surrogate subnetwork architecture}
\label{app:ablation_study}
In all experiments of the main paper, the surrogate subnetwork is implemented as a lightweight two-layer MLP with hidden dimension 128. Here, we further investigate how the surrogate subnetwork depth and hidden dimension affect the performance and efficiency of NDFA.

Table~\ref{tab:depth_dimension} summarizes the classification performance with different surrogate architectures on ConvNet and ResNet-32 over CIFAR-10 and CIFAR-100. We first examine the effect of the hidden dimension while fixing the surrogate depth to two layers. Increasing the hidden dimension from 128 to 512 leads to only marginal performance changes across architectures and datasets, suggesting that a relatively small hidden dimension is already sufficient for learning effective feedback. Therefore, we adopt the lightweight setting with hidden dimension 128 throughout the main experiments.

We next analyze the effect of surrogate subnetwork depth while fixing the hidden dimension to 128. Increasing the depth consistently improves classification accuracy on both architectures and datasets, indicating that deeper surrogate subnetworks can better approximate the nonlinear downstream transformations and provide more informative feedback signals. However, as shown in Table~\ref{tab:time}, increasing the depth also introduces substantially higher backward computation cost and reduces the training acceleration of NDFA. In particular, deeper surrogate subnetworks gradually diminish the parallel training advantage over BP.

These results suggest that lightweight MLP surrogate subnetworks provide a favorable balance between feedback quality and computational efficiency. Consequently, we employ a two-layer MLP with hidden dimension 128 in all experiments as an effective and efficient surrogate subnetwork.

\section{Surrogate subnetworks with layer-dependent capacities}
\label{app:layer_capacity}

We conduct experiments to investigate surrogate subnetworks with layer-dependent capacities. Specifically, we use AlexNet with six surrogate subnetworks on CIFAR-10 and evaluate a range of configurations in which the surrogate depth gradually increases or decreases across network layers. For each configuration, we report both classification accuracy and backward-pass time (values in parentheses denote the speedup relative to BP).

As shown in Table~\ref{tab:layer_capacity}, in most cases, assigning deeper surrogate subnetworks to earlier layers and shallower ones to later layers substantially improves accuracy, with only a modest reduction in training efficiency. In contrast, assigning shallower surrogate subnetworks to earlier layers and deeper ones to later layers not only increases the computational cost but also leads to lower accuracy. Note that for the highest-capacity setting (6→5→4→3→2→1), our method substantially improves accuracy compared with the uniform two-layer surrogate setting and achieves performance close to that of BP, while achieving a $1.10\times$ backward-pass speedup. The efficiency advantage becomes smaller as the surrogate depth approaches that of the corresponding BP path. The AlexNet architecture used in this experiment consists of eight layers. The first layer has seven downstream layers, whereas the associated surrogate subnetwork in this configuration has a depth of six. Therefore, the computational depth of the highest-capacity surrogate is already close to that of the corresponding BP path, which explains the limited backward-time improvement. 

These results suggest that allocating greater surrogate capacity to earlier layers is a more effective design choice for improving performance, despite a slight reduction in training efficiency. Nevertheless, our method with these settings remains much faster than BP due to its parallel training capability. In future work, we will investigate more principled approaches for designing layer-dependent surrogate subnetworks.

\begin{table}[H]
\centering
\caption{Results of surrogate subnetworks with layer-dependent capacities using AlexNet on CIFAR-10. The surrogate depths are listed from the earliest to the latest network layer.  Time is per-iteration backward time, and values in parentheses report speedup relative to BP.}
\label{tab:layer_capacity}
\footnotesize
\setlength{\tabcolsep}{2.0pt}
\renewcommand{\arraystretch}{1.12}
\begin{tabular}{lcccc}
\toprule
Depths (early $\to$ late) & Acc. ($\%$) & Time (ms) & FLOPs (G) & Params (M) \\
\midrule
$2\to2\to2\to2\to2\to2$ & $87.38\pm0.14$ & 0.92 (2.54$\times$) & 140.30 & 45.82 \\
$3\to3\to2\to2\to1\to1$ & $88.32\pm0.34$ & 1.63 (1.44$\times$) & 173.08 & 43.93 \\
$3\to3\to2\to2\to2\to2$ & $88.88\pm0.19$ & 1.69 (1.38$\times$) & 175.06 & 44.90 \\
$4\to4\to3\to3\to2\to2$ & $89.95\pm0.16$ & 1.90 (1.23$\times$) & 242.04 & 42.36 \\
$6\to5\to4\to3\to2\to1$ & $\mathbf{90.54\pm0.15}$ & 2.13 (1.10$\times$) & 277.15 & 39.65 \\
$2\to2\to2\to2\to3\to3$ & $87.43\pm0.20$ & 1.52 (1.54$\times$) & 209.56 & 79.39 \\
$1\to1\to2\to2\to3\to3$ & $87.03\pm0.15$ & 1.34 (1.75$\times$) & 202.62 & 76.00 \\
\bottomrule
\end{tabular}
\vspace{-5mm}
\end{table}

\section{Computational overhead}
\label{app:practical_efficiency}
We compare NDFA with BP in terms of per-iteration wall-clock time, FLOPs, and peak GPU memory on CIFAR-10 using ConvNet, AlexNet, and ResNet-32. The reported wall-clock time for NDFA does not account for parallel speedup. As shown in Table~\ref{tab:practical_cost}, NDFA introduces negligible overhead in FLOPs due to its lightweight surrogate subnetworks. Moreover, NDFA requires less GPU memory than BP because it does not need to store the end-to-end computational graph.

We further conduct a scaling analysis by progressively increasing the number of convolutional layers in ConvNet. The results are reported in Table~\ref{tab:practical_cost}, where CXF2 denotes a ConvNet with (X) convolutional layers and two fully connected layers, evaluated on CIFAR-10. As the network depth increases, the additional computational overhead introduced by NDFA remains negligible relative to BP, owing to the lightweight MLP-based surrogate subnetworks.

\begin{table}[H]
\centering
\caption{Comparison of computational overhead on CIFAR-10.}
\label{tab:practical_cost}
\small
\setlength{\tabcolsep}{7pt}
\begin{tabular}{lccc}
\toprule
Network & BP/NDFA Time (ms) & BP/NDFA FLOPs (G) & BP/NDFA GPU Memory (MB) \\
\midrule
C2F2 & 7.29 / 6.89 & 5.25 / 6.04 & 263 / 263 \\
C3F2 & 10.42 / 9.37 & 13.33 / 14.26 & 455 / 456 \\
C5F2 & 12.31 / 14.05 & 21.35 / 22.65 & 454 / 455 \\
C7F2 & 15.90 / 18.29 & 22.91 / 24.65 & 454 / 455 \\
AlexNet & 13.64 / 18.35 & 137.07 / 140.31 & 1894 / 1509 \\
ResNet-32 & 37.34 / 46.88 & 35.39 / 38.80 & 880 / 831 \\
\bottomrule
\end{tabular}
\end{table}

\clearpage

\newpage
\section*{NeurIPS Paper Checklist}

\begin{enumerate}

\item {\bf Claims}
    \item[] Question: Do the main claims made in the abstract and introduction accurately reflect the paper's contributions and scope?
    \item[] Answer: \answerYes{} 
    \item[] Justification: We do not claim any contribution out of scope. All the contributions mentioned are supported in the later sections, see Section~\ref{sec:Introduction}

\item {\bf Limitations}
    \item[] Question: Does the paper discuss the limitations of the work performed by the authors?
    \item[] Answer: \answerYes{} 
    \item[] Justification: We discuss the limitations of the work in Section~\ref{sec:Conclusion}. 

\item {\bf Theory assumptions and proofs}
    \item[] Question: For each theoretical result, does the paper provide the full set of assumptions and a complete (and correct) proof?
    \item[] Answer: \answerNA{} 
    \item[] Justification: Our work primarily focuses on the experimental perspective.

    \item {\bf Experimental result reproducibility}
    \item[] Question: Does the paper fully disclose all the information needed to reproduce the main experimental results of the paper to the extent that it affects the main claims and/or conclusions of the paper (regardless of whether the code and data are provided or not)?
    \item[] Answer: \answerYes{} 
    \item[] Justification: The  details to reproduce our experimental results are
described in the Appendix~\ref{app:hyperparams}

\item {\bf Open access to data and code}
    \item[] Question: Does the paper provide open access to the data and code, with sufficient instructions to faithfully reproduce the main experimental results, as described in supplemental material?
    \item[] Answer: \answerYes{} 
    \item[] Justification: : The code will be publicly available upon acceptance.

\item {\bf Experimental setting/details}
    \item[] Question: Does the paper specify all the training and test details (e.g., data splits, hyperparameters, how they were chosen, type of optimizer) necessary to understand the results?
    \item[] Answer: \answerYes{} 
    \item[] Justification: We specify all details in Appendix~\ref{app:hyperparams}

\item {\bf Experiment statistical significance}
    \item[] Question: Does the paper report error bars suitably and correctly defined or other appropriate information about the statistical significance of the experiments?
    \item[] Answer: \answerYes{} 
    \item[] Justification: We report the mean and standard deviation values by running each result for 3 times with different seeds.

\item {\bf Experiments compute resources}
    \item[] Question: For each experiment, does the paper provide sufficient information on the computer resources (type of compute workers, memory, time of execution) needed to reproduce the experiments?
    \item[] Answer: \answerYes{} 
    \item[] Justification: We provide the GPU type in the Appendix~\ref{app:hyperparams}
    
\item {\bf Code of ethics}
    \item[] Question: Does the research conducted in the paper conform, in every respect, with the NeurIPS Code of Ethics \url{https://neurips.cc/public/EthicsGuidelines}?
    \item[] Answer: \answerYes{} 
    \item[] Justification: We obey the NeurIPS Code of Ethics in every respect.

\item {\bf Broader impacts}
    \item[] Question: Does the paper discuss both potential positive societal impacts and negative societal impacts of the work performed?
    \item[] Answer: \answerNA{} 
    \item[] Justification: No societal impact is concerned by our work.
    
\item {\bf Safeguards}
    \item[] Question: Does the paper describe safeguards that have been put in place for responsible release of data or models that have a high risk for misuse (e.g., pre-trained language models, image generators, or scraped datasets)?
    \item[] Answer: \answerNA{} 
    \item[] Justification: Our work poses no such risks.

\item {\bf Licenses for existing assets}
    \item[] Question: Are the creators or original owners of assets (e.g., code, data, models), used in the paper, properly credited and are the license and terms of use explicitly mentioned and properly respected?
    \item[] Answer: \answerYes{} 
    \item[] Justification: We have cited all the papers whose datasets are used in our experiment.

\item {\bf New assets}
    \item[] Question: Are new assets introduced in the paper well documented and is the documentation provided alongside the assets?
    \item[] Answer: \answerNA{} 
    \item[] Justification: We do not release new assets.

\item {\bf Crowdsourcing and research with human subjects}
    \item[] Question: For crowdsourcing experiments and research with human subjects, does the paper include the full text of instructions given to participants and screenshots, if applicable, as well as details about compensation (if any)? 
    \item[] Answer: \answerNA{} 
    \item[] Justification: No experiment involves crowdsourcing or research with human subjects.

\item {\bf Institutional review board (IRB) approvals or equivalent for research with human subjects}
    \item[] Question: Does the paper describe potential risks incurred by study participants, whether such risks were disclosed to the subjects, and whether Institutional Review Board (IRB) approvals (or an equivalent approval/review based on the requirements of your country or institution) were obtained?
    \item[] Answer: \answerNA{} 
    \item[] Justification: No experiment involves crowdsourcing or research with human subjects.

\item {\bf Declaration of LLM usage}
    \item[] Question: Does the paper describe the usage of LLMs if it is an important, original, or non-standard component of the core methods in this research? Note that if the LLM is used only for writing, editing, or formatting purposes and does \emph{not} impact the core methodology, scientific rigor, or originality of the research, declaration is not required.
    \item[] Answer: \answerNA{} 
    \item[] Justification: We only use LLMs for writing.

\end{enumerate}
\end{document}